# Exploring the Use of Data-Driven Approaches for Anomaly Detection in the Internet of Things (IoT) Environment

Eleonora Achiluzzi, Menglu Li, Md Fahd Al Georgy, and Rasha Kashef

Toronto Metropolitan University

{eachiluzzi, menglu.li, mgeorgy, rkashef} @ryerson.ca

*Abstract*—The Internet of Things (IoT) is a system that connects physical computing devices, sensors, software, and other technologies. Data can be collected, transferred, and exchanged with other devices over the network without requiring human interactions. One challenge the development of IoT faces is the existence of anomaly data in the network. Therefore, research on anomaly detection in the IoT environment has become popular and necessary in recent years. This survey provides an overview to understand the current progress of the different anomaly detection algorithms and how they can be applied in the context of the Internet of Things. In this survey, we categorize the widely used anomaly detection machine learning and deep learning techniques in IoT into three types: clustering-based, classification-based, and deep learning-based. For each category, we introduce some state-of-the-art anomaly detection methods and evaluate the advantages and limitations of each technique.

## I. Introduction

The Internet of Things (IoT) is a rapidly expanding network that connects hardware, software, and devices through complex interconnections, enabling data collection and exchange [1]. As the number of IoT users and applications grows across various sectors, new challenges arise in the security and privacy of devices in the IoT network. One area of study in modern IoT data analytics is detecting anomalous data or outliers in data streams. Anomalous data, also known as anomalies or outliers, are unusual or unexpected patterns or behaviors in data that can indicate a problem or rare event. Errors or rare observations can cause anomalies. Errors can result from malicious attacks, compromising the entire IoT network if not detected and addressed. Rare observations are unusual events that may occur during the operation of the IoT network and may need to be monitored or alerted. Anomaly detection in the IoT is important for promptly identifying and addressing problems or unusual events. For example, anomalies in sensor data may indicate a malfunctioning device, while anomalies in network traffic data may indicate a cyber-attack. Anomaly detection can also identify unusual patterns in industrial processes or detect fraud or abnormal behavior in financial transactions.

Machine learning techniques have been widely used for anomaly detection in the IoT, as they can analyze large amounts of data efficiently and accurately. There are a variety of machine learning techniques that have been applied to anomaly detection in the IoT, including supervised learning techniques, such as support vector machines (SVMs) and decision trees, and unsupervised learning techniques, such as clustering algorithms and autoencoders [2]-[4]. These techniques can be applied to a wide range of data types, including sensor data, network traffic data, and financial transaction data. Identifying patterns in IoT data streams through various anomaly detection techniques can be useful for identifying malicious data, preventing network attacks, and detecting unusual operations in the IoT environment. Recently, several techniques have been commonly used for anomaly detection in IoT, such as statistical-based [2], proximity-based [3] or Machine Learning (ML)-based approaches and Deep Learning (DL)-based [4].

One challenge in using machine learning for anomaly detection in the IoT is the large amount of data generated by IoT devices. This data can be noisy, high-dimensional, and heterogeneous, making it difficult for machine learning models to detect anomalies accurately. Additionally, the dynamic nature of the IoT means that the data patterns and relationships may change over time, requiring the machine learning model to be re-trained or updated regularly. Another challenge in anomaly detection in the IoT is the limited availability of labeled training data. In many cases, obtaining a large enough dataset containing both normal and anomalous examples is challenging to train a machine-learning model effectively. This can be especially problematic when abnormal events are rare or hard to predict. Despite these challenges, machine learning techniques have shown promise for anomaly detection in the IoT. However, existing literature reviews on anomaly detection in the IoT using machine learning techniques often focus on a particular industrial application. For example, Ghosh et al. [4] primarily focus on outlier detection in wireless sensor networks, while Deorankar et al. [5] provide a survey on preventing cyberattacks. This leaves a research gap in providing a comprehensive overview of anomaly detection in the IoT and a list of state-of-the-art machine-learning techniques based on different algorithms.

This survey aims to address this research gap by outlining the current approaches to anomaly detection in IoT data and discussing the contributions and shortcomings of the presented works. This includes a review of the various machine learning techniques that have been applied to anomaly detection in the IoT, as well as a discussion of the challenges and limitations of these techniques. In this paper, we will focus on ML-based and DL-based techniques because of their ability to process and analyze a large amount of data. The techniques will be categorized into three types: clustering-based, classification-based, and deep learning-based. The state-of-the-art anomaly detection algorithms for each kind will be discussed.

The remainder of this paper is organized as follows: Section 2 introduces the background of anomaly detection in IoT applications using machine learning techniques. Section 3 presents the current clustering-based anomaly detection approaches. Section 4 highlights classification-based anomaly detection approaches on IoT. Section 5 presents selected deep learning-based anomaly detection approaches in the literature. Conclusions ad future directions are summarized in Section 6.

## II. BACKGROUND

### A. Anomaly Detection

Nonconforming patterns in big data are referred to as anomalies or outliers. The procedure of discovering patterns in a dataset whose behaviour is not what we predicted or expected is called anomaly detection. In general, anomaly refers to the outliers in a dataset, originating as part of the data cleansing process. However, anomaly or outlier detection can effectively detect irregular things in real-world problems, such as fraud detection, intrusion or damage detection, or even abnormal health condition identification [6]. Outliers can be classified into three categories: global outliers, contextual outliers, and collective outliers.

- Global outliers: A data instance can be considered a global outlier if it significantly differs concerning the remainder of the data set [7].

- Contextual outliers: A data instance is termed a contextual outlier if it is anomalous in a selected context but not otherwise [7].

- Collective outliers: This outlier refers to a collection of related data instances classified as anomalous with respect to the remainder of the data set [7].

Anomalous data in a computer network could indicate a hacked computer exposing sensitive data to an unauthorized user [8]. Anomalies in the IoT environment may occur for two major reasons: performance and security-related [8].

- Performance-related anomaly: This anomaly may occur due to network device malfunctions, such as sensor faults or unusual operations.

- Security-related anomaly: This anomaly may occur due to malicious IoT traffic attempting to obstruct and compromise a targeted system. Security-related anomalies can be organized into six categories: infection, exploding, probe, cheat, traverse, and concurrency [8]. Infection anomalies attempt to infect and damage a targeted system employing viruses or worms. Exploding anomalies attempt to deluge a system with bugs. A common example of an exploding anomaly is buffer overflow attacks. The third category, probe, attempts to collect information to expose the vulnerabilities of the targeted system, such as portmappers [8]. Cheat attacks are characterized by using abnormal callers, such as internet protocol or media access control spoofing [8]. The fourth category, traverse, attempts to match each possible key of a compromised system. A common example of a traverse anomaly is brute force attacks. Concurrency attacks exploit a targeted system by sending identical mass requests above the system's capacities [8]. Distributed Denial of Service (DDoS) and Botnet attacks are common examples of concurrency attacks.

Anomaly detection techniques can be utilized to detect and remove performance and security-related anomaly attacks in the IoT environment; however, anomaly detection presents challenges. Some common challenges in anomaly detection techniques include managing noise data and efficiently classifying normal behaviours from anomalous [9][10]. Noise may contribute to misinterpreting normal and anomalous data instances and must be removed before anomaly detection. The distinction between normal and abnormal data instances is often ambiguous and difficult to define.

### B. Machine learning for anomaly detection

Machine learning (ML), as a subset of artificial intelligence (AI), is a computer algorithm that can accomplish tasks without being explicitly programmed [11]. Machine learning builds a mathematical model based on the features of the historical data, and the model gets trained and updated when exposed to new sample data. After training, computer programs can learn, adjust actions, and make decisions automatically without human assistance. Recently, the number of applications of machine learning techniques has increased because they can help programmers implement computer programs faster, simpler, and more accurately [12]. Especially society produces overloaded data, and it is impossible to process and analyze all types of data by humans. Machine learning can process large amounts of data in a short time, and ML can even extract the features of relevant data in disordered datasets [12]. Another main advantage of machine learning is that ML keeps learning from the new training data and can improve and update itself if the algorithm produces unexpected outputs. Machine learning techniques are widely used to achieve classification, anomaly detection, and prediction tasks. Face/emotion recognition [13], sentiment analysis [14], and marketing recommendation systems [15] are well-known applications of ML in daily life. Because of the advantages of machine learning mentioned previously and the characteristic of anomaly detection, which involves analyzing and extracting the underlying patterns between a large amount of data, Machine learning is also one of the most popular techniques to detect anomalies. The ML techniques for anomaly detection can be categorized into three types: clustering-, classification-, and deep learning-based.

- Clustering-based algorithm: These algorithms apply the unsupervised learning technique, which uses unlabeled data to train the model. Therefore, each data point in the training set is unknown whether it is normal or abnormal. The main assumption for clustering-based algorithms to detect anomalies is that most of the instances in the dataset are normal, and normal data have some characteristics in common, so the anomaly points seem to be unfit for most of the dataset. The most widely used clustering-based anomaly detection techniques are K-mean clustering, Expectation-Maximization (EM), and FindCBLOF (Cluster-based Local Outlier Factor) [16].

- Classification-based algorithm: They apply the supervised learning technique, which requires each data

point to have a label as "normal" or "abnormal." The labelled data are passed to a classifier for training purposes. The classification-based algorithms operate anomaly detection under a general assumption also, which is that a classifier that can distinguish between normal points and anomalies can be learned in the given feature spaces [16]. Bayesian Networks and Support Vector Machines are examples of classification-based anomaly detection algorithms [16].

- Deep learning-based algorithm: These algorithms are based on artificial neural networks, both supervised and unsupervised learning. The neural network contains multiple layers to extract higher features from the input dataset. The widely used architectures for deep learning-based algorithms to detect outliers are Multi-Layer Perceptron (MLP), Convolutional Neural Networks (CNN), and Recurrent Neural Networks (RNN).

A summary of ML approaches for anomaly detection is presented in Table 1.

TABLE I. SUMMARY OF DIFFERENT TYPES OF MACHINE LEARNING TECHNIQUES FOR ANOMALY DETECTION

| ML Technique | Advantage | Disadvantage |
|---|---|---|
| Clustering-based | • Able to operate for unlabelled data.<br>• Easy to understand. | • Hardly adapt to dynamic network environments |
| Classification-based | • Supports multi-class identification, so that different types of anomalies can be distinguished.<br>• The testing phase is fast because the model has already been pre-computed. | • Unavailability of labelled data. |
| Deep learning-based | • Can extract high levels of underlying features between training data.<br>• RNN model can analyze data in real time. | • The computational complexity is high. |

## C. IoT

The Internet of Things (IoT) refers to the network of multipurpose devices connected to the internet. Data is collected from the devices, aggregated, processed, or stored, and exchanged between the devices [17]. A typical IoT system may include sensors, communication interfaces, advanced algorithms, and a cloud interface [17]. IoT applications are useful for various disciplines, such as energy, healthcare, education, and transportation. Some common applications for IoT devices include but are not limited to autonomous vehicles, smart contracts, and smart home appliances. The IoT allows conventional devices to become intelligent and autonomous [18]; however, the expanding IoT network imposes new challenges as billions of new devices are interconnected [17]. Identifying outliers in real-time IoT data analytics is essential for the security and privacy of IoT devices. The architecture of the Internet of Things contains four layers, as Fig. 1 indicates, including:

• Sensing Layer: it is integrated with hardware, such as sensors, to receive and record data [19][20].

• Networking Layer: This layer provides basic networking communication and data sharing over the wireless or wired network infrastructure [19][20].

• Service Layer: It creates and manages services, achieving applications and users' requirements [19][20].

• Interface Layer: This layer allows interaction between users and applications [19][20].

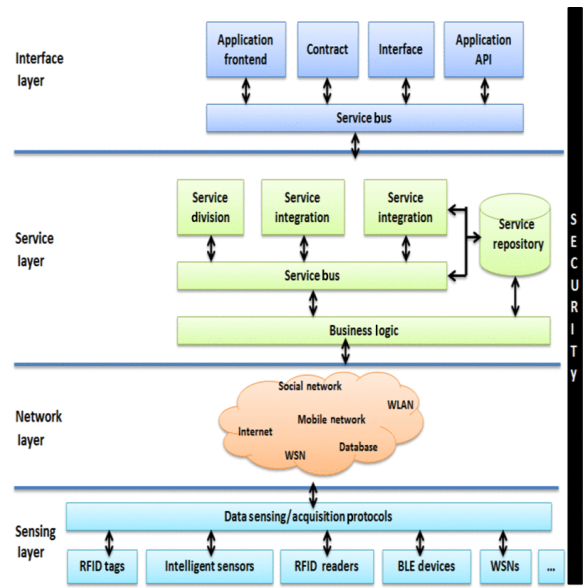

Figure 1. The multi-layer architecture of IoT [19]

## III. CLUSTERING-BASED ANOMALY DETECTION

Clustering is an unsupervised machine learning approach that aims to group input datasets into clusters, such that the objects in a cluster are similar to one another and dissimilar from objects in a separate cluster. Clustering-based methods are widely used for anomaly detection in IoT applications. Many works have been conducted to prove the effectiveness and accuracy of integrating clustering techniques to detect outliers in an input dataset. For instance, Alguliyev et al. [21] proposed a hybrid model to detect anomalies accurately in big data by combining particle swarm optimization (PSO) and K-means algorithms. This work presents a novel weighted clustering method, such that inter-cluster distances are maximized, and intra-cluster distances are minimized [21]. The proposed method eliminates the presence of predefined cluster centers and many local minimum points. The proposed method is only tested using one dataset, and experimental results prove that it is superior to the traditional K-means algorithm in performance and clustering accuracy. Clustering algorithms typically require an inputted set of parameters by the user. K-means clustering assumes clusters of spherical shapes; therefore, Rahman et al. [22] present a novel density-based clustering algorithm for outlier detection that is parameter independent and can cluster data of arbitrary shapes. The proposed parameter-independent density-based clustering (PIDC) algorithm works in two stages. The first stage identifies outliers in the datasets using unique closest neighbour concepts and removes the detected outliers. Then, the records are passed into the second stage for density-based clustering. This work is tested using six datasets, and a comparative analysis is conducted to evaluate the performance of the proposed algorithm to five common clustering methods. The proposed algorithm performs well on high-dimensional datasets with outliers; however, it exhibits high computational complexity.

Anomaly detection in IoT applications often includes real-time data requiring prompt outlier detection. The fast arrival of data requires fast computation with minimal memory usage [23]. In this context, Bah et al. [23] present a novel hybrid model called Micro-Cluster with Minimal Probing (MCMP) by combining Micro-Cluster Outlier Detection (MCOD) and Thresh_LEAP methods to address these challenges. This work aims to minimize the computational speed and memory consumption while detecting distance-based outliers accurately. Micro-clusters store neighbouring data points to minimize distance-based computations [23], improving time and memory consumption. The principle of minimal probing from Thresh_LEAP was used for objects outside the micro-clusters and improved to compute the significance of inliers. When the proposed algorithm is applied to large real-world and synthetic datasets, it minimizes computational cost and memory consumption; however, when dealing with objects outside the micro-clusters, their approach misidentifies some crucial outliers.

Cluster-based Local Outlier Factor (LOF) algorithm effectively detects local outliers, however, computing the LOF at each data point adds computational overhead. In addition, if the K-distance neighbourhood surrounding an outlier contains outliers misidentified as normal points, then the outlier may be incorrectly identified as normal. Yang et al. [24] present a Neighbour Entropy Local Outlier Factor (NELOF) outlier detection algorithm, utilizing an improved Self-Organizing Feature Map (SOFM) algorithm to cluster the dataset. The improved SOFM algorithm uses the Canopy algorithm to avoid the random selection of neurons and adjusts the neurons dynamically. The improved SOFM algorithm proves superior to the traditional SOFM algorithm, significantly decreasing the number of dead neurons. Furthermore, this work replaces the K-distance neighbourhood with relative K-distance neighbourhoods to reduce the influence of outliers in K-distance neighbourhoods [24]. The proposed NELOF algorithm was tested with seven different datasets and proved to outperform the LOF algorithm in execution time and accuracy; however, this work does not explore the effectiveness of the proposed algorithm on high-dimensional datasets.

Since anomaly detection is critical for time-sensitive IoT applications that include high-dimensional data, Lyu et al. [25] present a Fog-Empowered anomaly detection method using hyperellipsoidal clustering (HyCARCE). Fog computing is presented as an alternative to Cloud computing to provide off-loading for the Cloud [25]. In the Fog architecture, the end nodes or sensors transmit data directly to the Fog nodes for clustering and anomaly detection, improving detection response and time delay by minimizing the computational overhead at the sensors and the Cloud. The advantages of using HyCARCE for data clustering include its ability to automatically select the number of clusters and its accommodating to different data distributions, such as linear or hyperspherical. The proposed method is applied to two real-world and two synthetic datasets, and it is proven to detect outliers and anomalous clusters in the dataset accurately and in a timely manner. This work does not address privacy and security concerns due to the exchange of information in the Fog architecture.

*A. Summary of selected algorithms*

A summary of selected clustering-based approaches for anomaly detection is presented in Table 2.

TABLE II.    SUMMARY OF SELECTED CLUSTERING-BASED ANOMALY DETECTION APPROACH ON IoT

| Model & Year | Contributions | Limitations |
|---|---|---|
| PSO & K-means (2019) [21] | • Combines PSO and K-means algorithms to develop a novel weighted clustering method for anomaly detection.<br>• Their method eliminates the presence of predefined cluster centers and multiple minimum points while improving accuracy. | • The proposed method is only tested for one dataset.<br>• The performance of the proposed method is only compared to the traditional K-means algorithm. |
| PIDC (2018) [22] | • A novel density-based clustering algorithm which is parameter independent can detect clusters of arbitrary shapes. | • High computational complexity. |
| MCMP (2019) [23] | • Proposes a novel MCMP approach reducing computational speed and memory consumption during outlier detection.<br>• Able to compute the significance of inliers. | • The proposed approach misidentifies some crucial outliers when dealing with objects outside the micro-clusters. |
| NELOF & SOFM clustering algorithm (2019) [24] | • An improved SOFM clustering algorithm that utilizes the Canopy algorithm to avoid the random selection of neurons.<br>• A relative K-distance neighbourhood to reduce the influence of outliers that exist in the K-distance neighbourhood. | • The performance of the proposed method is only compared to LOF in the analysis.<br>• Does not explore the effectiveness of the proposed algorithm on high-dimensional datasets. |
| HyCARCE (2017) [25] | • A novel Fog-Empowered detection using HyCARCE.<br>• Sensor data is sent to the Fog nodes to reduce detection time<br>• Supports location-awareness of anomalies. | • Does not address privacy and security concerns from information exchange in the Fog nodes. |

## IV. CLASSIFICATION-BASED ANOMALY DETECTION

In machine learning, there are many techniques that we can use to classify things. In a typical classification-based problem, there is a balanced number of positives and negatives in the dataset. So, we train the model on a balanced number of positives and negatives. In anomaly detection problems, there are fewer positive examples than negative ones. The positive examples (anomalous data) might be lesser than 5% of the total data. The main task of classification is to identify the category or class label of new instances from a data set. Classifying the algorithm, called a classifier, depends on the learning from a training dataset. The training dataset contains data that is correctly classified into accurate class labels. Similarly, for anomaly detection, classification algorithms will try to classify data into two broad categories – normal and abnormal. Some common classification techniques for detecting anomalies are discussed below:

- Classification Tree: It is like a tree pattern graph, also known as a prediction model or decision tree. The internal nodes are called test properties, each branch signifies the test results, and the final leaves indicate the class to which the test data belongs. The two most used algorithms are ID3 and C4.5 [26]. Classification tree approaches, when compared to naïve Bayes classification, the result obtained from decision trees was found to be more accurate [27].

- Support Vector Machine (SVM): SVM is popular in recognizing patterns and is widely used in intrusion detection systems. "When compared to neural networks in the KDD cup data set, it was found that SVM outperformed NN in terms of false alarm rate and accuracy in most attacks" [28].

- Naïve Bayes network: To take benefit of the structural connection between the random variables, we can use a probabilistic graph model - Naïve Bayesian Networks. The model delivers a solution to the question if only a limited number of observed events are known, then what is the likelihood of a specific kind of attack. Let's compare the decision tree and Bayesian techniques, although the precision of the decision tree is far better. The computational time of the Bayesian network is found to be low [27].

- Genetic Algorithm: Genetic Algorithm (GA) was first introduced in the computational biology field. However, the GA is applied for intrusion detection to develop a set of classification rubrics from the network audit data. The substantial properties of GA are its strength against noise and self-learning competencies [29].

- HADES- IoT is a novel host-based anomaly detection and prevention approach. The model is based on whitelisting legitimate processes on an IoT device. The notion of this method is that simple programs that are recognized to run on an "uninfected" off-the-shelf device are allowed to run. To build a whitelist of benign programs, profiling must be performed once for each device. The devices with enabled Telnet service by default (e.g., SimpleHome IP camera) are potentially vulnerable to Mirai" [30]. In the proposed model (HADES-IoT), even such a default misconfiguration does not cause harm since the execution of any unauthorized binary is terminated upon its spawning.

### A. Summary of selected algorithms

A summary of selected classification-based approaches for anomaly detection is presented in Table 3.

## V. DEEP LEARNING-BASED ANOMALY DETECTION

Deep learning, based on artificial neural networks to process data, has demonstrated its ability to extract features and high accuracy for classification tasks. Therefore, applying deep learning-based algorithms on the Internet of Things environment to detect anomalies is also popular. Several selected deep learning-based algorithms are elaborated on in the following literature. One of the common examples of detecting anomalies in the IoT is detecting network attacks. Some network intrusion detection systems (NIDS) have problems with adaptation to dynamic network environments, unavailability of labelled data, and high false-positive rates. Therefore, Van et al. [36] proposed using deep learning algorithms to implement network intrusion detection systems. Two types of deep learning models are mentioned in their paper: the Restricted Boltzmann Machines (RBM) and Autoencoder (AE). The authors constructed a stacked RBM and AE as two Deep Belief Network (DBN) structures and compared their performance on detection intrusion. For the stacked RBM, which uses a probability distribution, the hidden layer of each RBM is set to be the input layer of the next RBM of the stack. The stacked AE can extract features of network data by unsupervised learning so that it can solve the challenge of the unavailability of labelled data. For the experiment, Van et al. [36] used the same dataset that contains four types of network attacks to test the ability of two DBM models. The results show the Stacked AE outperforms the Stacked RBM on the accuracy of intrusion detection; however, the training time and execution time of the stacked RBM are much longer than the Stacked AE model. Almiani et al. [37] proposed a deep learning-based intrusion detection system in the IoT environment. Their model contains two major components: the traffic analysis engine and the classification engine. The traffic analysis engine is used to pre-process the traffic data, such as symbolic-to-numeric transformation, feature reduction and normalization. Then, the processed data is fed into the classification engine, which adopts two deep recurrent neural networks (RNN) to respond fast in a real-time environment. The two RNN work as two filters of attack detection. The traffic data classified as normal by the first RNN layer will be passed to the second RNN detection layer to identify whether it is an anomaly. The same dataset is used to train the two layers of RNN. The only difference is the training set for the first RNN contains both normal and abnormal data, while the training set for the second RNN only has normal traffic data. Almiani et al. [37] compare their proposed model with other baseline models for anomaly detection accuracy and execution time, which indicates the proposed RNN model has a high sensitivity to detect abnormal attacks in a competitive computational overhead. The scalability and distribution of resources are the characteristics of the Internet of Things. Therefore, any anomaly detection model that depends on a centralized cloud will fail to handle the IoT requirements [38].

TABLE III. SUMMARY OF SELECTED CLASSIFICATION-BASED ANOMALY DETECTION APPROACH ON IoT

| Model & Year | Contributions | Limitations |
|---|---|---|
| Naive Bayes + decision Tree (2010) [31] | • The model performs balance detections and keeps false positives at an acceptable limit for different network attacks.<br>• The method minimized the rate of false positives and maximized balanced detection rates. [28] | • The model requires much improvement in detecting false positives to remote-to-user attacks. |
| HADES-IOT (2019) [30] | • The generic model can be easily adapted to any Linux kernel version. The method is said to be resilient against attackers and focused on disabling its protection mechanisms.<br>• HADES-IoT terminates any executable not included in the whitelist both of them upon its execution and stops the attack. | • Researchers initially utilized features provided by Linux, such as KProbe1 or inotify. After examining several IoT devices (D-Link IP Camera, SimpleHome IP camera), they found that these features are not supported.<br>• As the challenge is to distinguish unidentified processes in real-time right after their spawning, the Linux process scheduler is another limiting factor.<br>• The model is based on the whitelisting approach. This may be viewed as impractical because some benign programs may be missed during profiling. |
| One Class and Two Class SVM (2012) [32] | • The first-class SVM is used for detecting abnormality scores. Secondly, the detector is retrained when certain new data records are included in the existing dataset.<br>• A prior failure history is not required, and the model continually learns from the observed failure events. | • It has no performance measure of computational complexity. |
| DT + SVM (2007) [33] | • The data set is first passed through the tree, and node information is generated and passed along with the original set of attributes through SVM to obtain the final output. [30] | • This approach delivers equivalent results to SVM. |
| Ensemble approach [33] | • To make the final decision, information is combined from different individual classifiers.<br>• Provided the best performance for Probe and R2L classes. | • Selecting base classifiers is critical and cannot be done automatically. |
| k-Means +ID3 (2010) [34] | • Proposed a hybrid technique which combines clustering and classification.<br>• The K-mean clustering can be applied to the normal training data points to form clusters. Then the decision tree will be performed on each cluster.<br>• The proposed algorithm outperforms the individual k-Means and the ID3 method. | • This approach is limited to a specific dataset. |
| k-Medoids +Naïve Bayes (2012) [35] | • Similar data instances are grouped by using the k- Medoids clustering technique.<br>• It can increase detection accuracy rate and reduction in false alarm rates. | • It is challenging to predict naïve Bayes classifiers in different settings. |
| SVM + k-medoids (2012)[35] | • Proposed a hybrid technique which combines k-medoids clustering and SVM classification.<br>• The proposed anomaly detection technique reaches a higher accuracy than the baseline algorithms. [32] | • Time complexity is more when the dataset is very large. |

N.g. et al. [38] proposed Vector Convolutional Deep Learning (VCDL) model to detect anomalies in IoT traffic, which applies an emerging distributed intelligence approach called "fog computing." The proposed model contains three layers of components. The first layer is IoT devices which are distributed. The second layer is the fog layer—multiple work fog nodes connected to the IoT devices and train each VCDL model in a distributed manner. The master fog node in the fog layer will collect and share the best set of parameters with the worker nodes. Therefore, the traffic data will be fed into the corresponding worker node and be classified as either normal or attack. The classification result will be passed to the cloud layer, the third layer of the proposed framework. The cloud layer is used to validate the information from the entire fog layer. The experiment results indicate that the proposed distributed VCDL framework can detect anomaly traffic data with high accuracy and less detection time than the centralized detection model.

The IoT system always involves real-time data or time-series data. Therefore, several pieces of research focus on detecting anomalies in time series data collected by IoT devices. For example, Liu et al. [39] worked on detecting outlier data in the indoor climate control system. There are two kinds of anomalies: point anomaly, which indicates a single outlier value hugely different from other data, and contextual anomaly, which refers to a series of inappropriate data points. Liu et al. [39] proposed a neural network-based model to detect these two kinds of anomalies, which combines the autoencoder (AE) and the long short-term memory (LSTM) model. The structure of AE is similar to the regular Feed Forward Neural Network with a reduced number of neurons in hidden layers so that the output of the AE can be very close to the input. The LSTM model can extract features of sequential data and capture the relationship between neighbouring input data [39]. Therefore, the AE component in the proposed model works to detect point anomalies, while the LSTM component detects the contextual anomaly. The proposed model demonstrated the accuracy of the anomaly detection algorithm could be improved by integrating the neural network models.

## A. Summary of selected algorithms

A summary of selected deep learning-based approaches for anomaly detection on IoT is presented in Table 4, which includes the contributions and drawbacks of each algorithm.

## VI. CONCLUSION AND FUTURE DIRECTIONS

The Internet of Things (IoT) has recently become popular as a system connecting physical hardware, such as computing devices, sensors, and other technologies. Over the IoT network, data can be collected, transferred, and exchanged with other devices without requiring human interactions, and it always involves processing and analyzing a massive amount of data. Therefore, the security and operation status of the IoT network is important, and it is essential to detect any unusual behaviours known as anomalies. The anomaly in the IoT environment can be categorized as performance-related or security-related. The performance-related anomaly can be represented as sensor faults or unusual events. The security-related anomaly is usually caused by malicious attacks. In this paper, we provided a survey on the state-of-the-art anomaly detection algorithms in the IoT environment using machine learning techniques because machine learning technology has a strong ability to deal with the big data involved in the IoT. We categorize the machine learning-related algorithms to detect anomalies into three types: clustering-based, classification-based, and deep learning-based. For each detection algorithm, we select a significant number of research papers, discuss the advantages of the chosen method, and mention each existing method's limitations. We also notice that although much work has been done through independent algorithms, deploying hybrid approaches can provide better results and overcome the shortcoming of one method over the other. We believe that this survey paper provides an in-depth knowledge of these commonly used approaches and will help make decisions on choosing a particular technique for an anomaly detection problem in the IoT environment. Future direction involves the use of multi-level learning [41], semantic learning [42], ensemble learning [43], hybrid learning [44], and distributed computing [45]-[46] for anomaly detection.

TABLE IV. SUMMARY OF SELECTED DEEP LEARNING-BASED ANOMALY DETECTION APPROACH ON IoT

| Model & Year | Contributions | Limitations |
| --- | --- | --- |
| DBN (2017) [36] | • Gives two examples of deep learning-based models for anomaly detection.<br>• Provides comparisons in anomaly detection accuracy and computational performance for the two selected models. | • Does not choose a non-deep learning-based anomaly detection model as the baseline to demonstrate the ability of deep learning.<br>• The types of intrusions that the proposed models can detect are limited. |
| 2-RNN (2020) [37] | • Proposed a two-layer RNN model to increase the anomaly detection accuracy on a challenging dataset.<br>• The proposed model can effectively detect in real-time environments with a competitive computational overhead. | • Only apply a sample dataset in the experiment to test the effectiveness of the proposed model. |
| VCDL (2020) [38] | • Trains the Vector Convolutional Deep Learning model in a distributed manner.<br>• Able to detect the anomaly traffic in parallel.<br>• Achieve detection performance with high accuracy. | • Only applies a sample dataset in the experiment to test the effectiveness of the proposed model.<br>• The model does not work for detecting multiclass anomalies. |
| AE-LSTM (2020) [39] | • Integrates two types of neural network models to detect different types of anomalies.<br>• The proposed anomaly detection algorithm can be performed in a real-time situation. | • They may not be able to detect anomalies with different patterns.<br>• It has no performance measure of computational complexity. |
| CNN-based monitoring detection (2019) [40] | • Proposed two CNN-based models to detect two types of abnormal situations in the operating environment of power equipment.<br>• It outperforms the traditional CNN network for detecting personnel or fire smoke in images. | • The purpose of the anomaly detection algorithm is very specific. The model cannot detect other abnormal cases in the operating environment.<br>• In the experiment, only images are used for testing. The ability of anomaly detection on video is not proven. |